\documentclass{article}

\usepackage[final]{neurips_2022_ml4ps}

\usepackage[utf8]{inputenc} 
\usepackage[T1]{fontenc}    
\usepackage{afterpage}
\usepackage{hyperref}       
\usepackage{url}            
\usepackage{booktabs}       
\usepackage{amsfonts}       
\usepackage{nicefrac}       
\usepackage{microtype}      
\usepackage[svgnames,table]{xcolor}
\usepackage{graphicx}
\usepackage{caption}
\usepackage{subcaption}
\usepackage{enumitem}      
\usepackage{pifont}

\graphicspath{ {./figures/} }

\usepackage{amsmath,amsfonts,bm}

\def\eqref#1{equation~\ref{#1}}

\def\1{\bm{1}}

\DeclareMathAlphabet{\mathsfit}{\encodingdefault}{\sfdefault}{m}{sl}
\SetMathAlphabet{\mathsfit}{bold}{\encodingdefault}{\sfdefault}{bx}{n}

\def\gF{{\mathcal{F}}}

\def\gM{{\mathcal{M}}}

\def\gT{{\mathcal{T}}}

\newcommand{\E}{\mathbb{E}}

\newcommand{\R}{\mathbb{R}}

\newcommand{\Var}{\mathrm{Var}}

\DeclareMathOperator*{\argmin}{arg\,min}

\newcommand\norm[1]{\lVert#1\rVert}
\newcommand{\dd}[2]{\frac{d#1}{d#2}}
\newcommand{\pdd}[2]{\frac{\partial#1}{\partial#2}}   

\title{Learning Integrable Dynamics with Action-Angle Networks}

\author{%
  Ameya Daigavane\thanks{Work performed when affiliated with Google Research and visiting Flatiron Institute.} \\
  Massachusetts Institute of Technology \\
  Cambridge, MA 02139, USA \\
  \\
  \texttt{ameyad@mit.edu} \\ 
  \And
  Arthur Kosmala \\
  Ludwig-Maximilians-Universität München \\
  80333 Munich, Germany\\
  \\
  \texttt{a.kosmala@physik.uni-muenchen.de} \\
  \And
  Miles Cranmer \\
  Princeton University\\
  Princeton, NJ 08544, USA \\
  \\
  \texttt{mcranmer@princeton.edu} \\
  \And
  Tess Smidt \\
  Massachusetts Institute of Technology \\
  Cambridge, MA 02139, USA \\
  \\
  \texttt{tsmidt@mit.edu} \\ 
  \And
  Shirley Ho \\
  Centre for Computation Astrophysics, Flatiron Institute \\
  New York, NY 10010, USA \\
  \\
  \texttt{shirleyho@flatironinstitute.org}
}

\begin{document}

\maketitle


\begin{abstract}
Machine learning has become increasingly popular for efficiently modelling the
dynamics of complex physical systems,
demonstrating a capability to learn effective models for dynamics which ignore redundant degrees of freedom.
Learned simulators typically predict the evolution of the system in a step-by-step manner with numerical integration techniques. 
However, such models often suffer from instability over long roll-outs
due to the accumulation of both estimation and integration error
at each prediction step. Here, we propose an alternative construction
for learned physical simulators that are inspired by the concept of \emph{action-angle coordinates}
from classical mechanics for describing integrable systems.
We propose \emph{Action-Angle Networks}, which
learn a nonlinear transformation from
input coordinates to the action-angle space,
where evolution of the system is linear.
Unlike traditional learned simulators,
Action-Angle Networks do not employ any higher-order numerical integration methods, making them extremely efficient at modelling the dynamics of integrable physical systems.
\end{abstract}


\section{Introduction}
\label{sec:intro}

\subsection{Modelling Hamiltonian Systems}

Hamiltonian systems are an important class of physical systems whose dynamics are governed by a scalar function $H$, called the Hamiltonian. The state of a Hamiltonian system is a 2-tuple $u(t) = (q(t), p(t))$, where
$q \in \mathbb{R}^n$ are the positions and  $p \in \mathbb{R}^n$ are the canonical momenta.
We are interested in predicting the
time evolution of a Hamiltonian system as a function of time $t$.
In particular, we seek to learn a model $\gM_\phi$ (parameterized by $\phi$) that can accurately predict the future state 
$u(t + \Delta t)$ given the current state $u(t)$:
\vspace{-0.2em}
\begin{align}
    \gM_\phi(u(t), \Delta t) \approx u(t + \Delta t).
\end{align}
Previous efforts
towards modelling Hamiltonian systems \citep{hnn,lnn,sympnets,chen:node,kidger:nde} have seen success with neural networks trained via backpropagation on the trajectory prediction objective:
\begin{align*}
    \phi^* = \argmin_\phi \ \sum_{t}
    \norm{\gM_\phi(u(t), \Delta t) - u(t + \Delta t))}^2.
\end{align*}
Such models are often constrained in some way to  match the underlying physical evolution, which improves their accuracy. However, they suffer from several drawbacks that have restricted their effectiveness: \textbf{(1)} Their predictions tend to be unstable over long roll-outs (when $\Delta t$ is large). \textbf{(2)} They tend to require many parameters and need long training times. Further, the inference times of these models often scale with $\Delta t$.

Based on the principle of \emph{action-angle coordinates} \citep{arnol'd}
from classical mechanics, we
introduce a new paradigm for learning physical simulators which incorporate an inductive bias for learning \emph{integrable} dynamics.
Our Action-Angle Network 
learns an invertible transformation to action-angle
coordinates
to linearize the dynamics in this space. In this sense, the Action-Angle Network can be seen as a physics-informed adaptation of DeepKoopman \citep{lusch:koopman}.
The Action-Angle Network 
is efficient both in parameter count and inference cost; it
requires much fewer
parameters than previous
methods to reach similar performance and enjoys
an inference time independent of $\Delta t$.

\subsection{Action-Angle Coordinates}
The time evolution of any Hamiltonian system is given by Hamilton's equations:
\vspace{-0.2em}
\begin{align}
        \dd{q}{t} = \pdd{H}{p}, \quad \dd{p}{t} = -\pdd{H}{q}
        \vspace{-0.5em}
\end{align}
where $H$ is the Hamiltonian of the system. The complexity for learning these dynamics arises solely because $H$ is not known and must be inferred; only samples from a trajectory over which $H$ is conserved is available.
The core issue for all physical simulators is the non-linearity of the dynamics
when expressed in the canonical coordinates $(q, p)$. 
However, for \emph{integrable} systems
\citep{tong:dynamics} which possess significant symmetry,
the dynamics are actually linear in a set of coordinates termed the \emph{action-angle}
coordinates \citep{arnol'd}.
The actions $I$ and angles $\theta$ are related to the canonical coordinates $(q, p)$ via
a symplectic transformation $\gT$ \citep{symplectic:meyer2013introduction} as $(I, \theta) = \gT(q, p)$. In the action-angle coordinates, the Hamiltonian only depends on the actions $I$,
not the angles $\theta$. Thus, Hamilton's equations in this basis tell us that:
\vspace{-0.2em}
\begin{align}
        \dd{I}{t} = -\pdd{H}{\theta} = 0, \quad \dd{\theta}{t} = \pdd{H}{I}.
    \label{eqn:theta-dot}
\end{align}
Thus, the actions $I$ are always constant across a trajectory,
while the angles evolve linearly with constant rate $\dot{\theta} = \frac{\partial H}{\partial I}$.
The action-angle space can be thought of as a torus $\mathbb{T}^n$,
where the actions $I$ are a function of the radii, and the angles
describe the individual phases living in $[0, 2\pi)$.


Learning the mapping $\gT$ from canonical coordinates $(q, p)$ to action-angle coordinates $(I, \theta)$ for several physical systems
was first explored in \citep{learning-symmetries}, which did not focus on the complete dynamics of integrable systems.
We leverage their framework to additionally incorporate a dynamics model $\gF$ to learn a complete physical simulator.

\cite{ishikawa:integrable} attempt to model the time evolution of actions for integrable systems by learning the Hamiltonian in action-angle coordinates. However, their overall objective and training setup are different from ours: they have samples of the true action-angle coordinates $(I, \theta)$ as input, while we only observe the canonical coordinates $(q, p)$.


\section{Action-Angle Networks}
\label{sec:method}

\begin{figure}[htbp]
    \centering
    \includegraphics[width=0.78\textwidth]{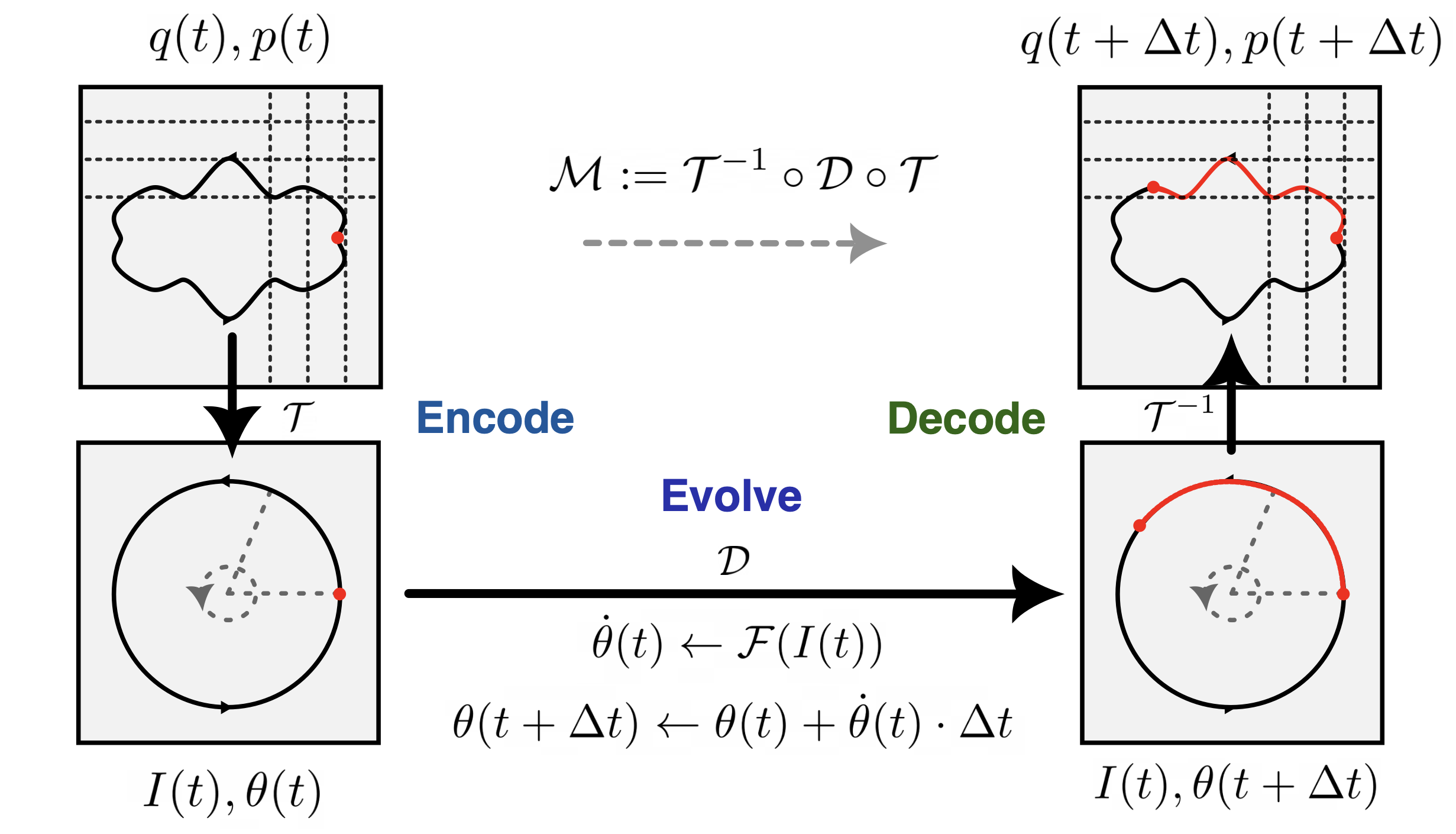}
    \caption{An overview of the Action-Angle Network.}
    \label{fig:model_diagram}
\end{figure}

Suppose we observe the system at time $t$, and we query our model $\gM_\phi$ to predict the state at a future time $t + \Delta t$. The Action-Angle Network performs the following operations, illustrated in \autoref{fig:model_diagram}:
\begin{itemize}[leftmargin=*]
    \item \textbf{Encode}: Convert the current state in canonical coordinates $q(t), p(t)$ to action-angle coordinates $I(t), \theta(t)$ via a learned map $\gT$.
    \item \textbf{Evolve}: Compute the angular velocities $\dot{\theta}(t)$ at this instant with the dynamics model $\gF$: \\$\dot{\theta}(t) \gets \gF(I(t))$.
    Evolve the angles to time $t + \Delta t$ with these angular velocities:
    \begin{align}
        \theta(t + \Delta t) \gets \theta(t) + \dot{\theta}(t) \cdot \Delta t \pmod{2\pi} 
    \end{align}
    \vspace{-1.5em}
    \item \textbf{Decode}: Convert the new state back to canonical coordinates $q(t + \Delta t), p(t + \Delta t)$ via $\gT^{-1}$.
\end{itemize}


\textbf{Symplectic Normalizing Flows}:
The transformation $\gT$ from canonical coordinates to action-angle coordinates is guaranteed to be symplectic \citep{arnol'd}.
To constrain our encoder to learn only symplectic transformations, we 
compose layers of
symplectic normalizing
flows \citep{sympflow,learning-symmetries,sympnets}.
In particular, we found that the G-SympNet layers \citep{sympnets}
with our modification described below
performed well empirically. G-SympNet was shown to be universal;
they can represent any symplectic transformation given sufficient width and depth.
Similar to affine coupling layers \citep{nice},
each G-SympNet layer $\psi_i$ operates on only one of the coordinates
keeping the other fixed, depending on the parity of $i$:
\begin{align}
    \psi_{2k}(q, p) = (q, p + f(q)), \quad
    \psi_{2k + 1}(q, p) = (q + f(p), p)
\end{align}
where $f$ is of the following form $f(x) = Cx + W^T \text{diag}(A)\sigma(Wx + B)$ with learnable parameters $A \in \R^{d_o}, B \in \R^{d_o}, C \in \R$
and $W \in \R^{d_o \times n}$, where $d_o$ is a hyperparameter.
Our minor modification above allows each layer to model the identity transformation,  enabling the training of deeper models.

\textbf{Action-Angles via Polar Coordinates}:
We found that learning a mapping from canonical coordinates which live in $\mathbb{R}^{2n}$
directly to action-angle coordinates which live in the torus $\mathbb{T}^n$ was challenging for the network.
We hypothesize that the differing topology of these spaces is a major obstacle because of the resulting singularities.
To bypass this, we borrow a trick from \cite{learning-symmetries};
we have the G-SympNet instead output the components of the actions $(I^{(x)}, I^{(y)})$ in the Cartesian coordinate basis.
These are then converted to action-angles $(I, \theta)$ in the polar coordinate basis, by the standard transformation $\gT_{\text{polar}}$
applied to each pair of $(I^{(x)}_i, I^{(y)}_i)$ coordinates:
\begin{align}
    I_i = \sqrt{(I^{(x)}_i)^2 + (I^{(y)}_i)^2}, \quad 
    \theta_i = \arctan(I^{(y)}_i / I^{(x)}_i)
\end{align}
Our encoder can thus be described as $\gT := \gT_\text{Polar} \circ \text{G-SympNet}$.

\textbf{Evolve}: From \autoref{eqn:theta-dot}, we know that the angular velocities $\dot{\theta}$ are only a function of the actions $I$
at the instant $t_0$, which we model as a simple multi-layer perceptron (MLP) $\gF$.
Since the angles $\theta$ are guaranteed to evolve linearly,
we do not need to use higher-order numerical integration schemes.
Instead, a single call to the forward Euler method is exact:
\begin{align}
    \dot{\theta}(t_0) &\gets \gF(I(t_0)) \quad 
    \theta(t_0 + \Delta t) \gets \theta(t_0) + \dot{\theta}(t_0) \cdot \Delta t \pmod{2\pi}
\end{align}
\textbf{Decode}: As $\gT$ is symplectic, $\gT$ is invertible. 
Thus, our decoder is $\gT^{-1} := \text{G-SympNet}^{-1} \circ  \gT_{\text{polar}}^{-1}$.

\subsection{Training}
We generate a trajectory $\text{Tr} = \{(q(t), p(t))\}_{t = 1}^T$
of $T = 1000$ time steps, by either evaluating the closed-form dynamics equations or via numerical integration.
In practise, this trajectory can also be obtained from raw observations, and does not need to be regularly sampled in time. We train models upto the first $500$ time steps and evaluate performance on the last $500$ time steps, for each trajectory. 

\textbf{State Prediction Error}: We wish to learn parameters $\phi := (\phi_{\gT}, \phi_{\gF})$ that minimize the prediction loss $L_{\text{predict}}$:
\vspace{-0.8em}
\begin{align}
   L_{\text{predict}} = \textstyle\frac{1}{1 + \Delta t}
   \textstyle\sum_{t_0 \in \text{Tr}} \norm{\gM_\phi(u(t_0), \Delta t) - u(t_0 + \Delta t)}^2.
\end{align}
\textbf{Regularization of Predicted Actions}:
Additionally, we wish to enforce the fact that the predicted actions $I$ are constant across
the trajectory. We do this by adding a regularizer $L_{\text{action}}$ on the variance of the predicted actions:
\vspace{-0.3em}
\begin{align}
    L_{\text{action}} &= \textstyle\widehat{\Var}(I) = \frac{1}{T}\sum_{t_0 \in \text{Tr}} \left(I(t_0) - \widehat{\E}[I]\right)^2
    \text{where} \
    \widehat{\E}[I] = \frac{1}{T}\sum_{t_0 \in \text{Tr}}I(t_0).
\end{align}
We find that this global regularizer works better than the loss proposed in \cite{learning-symmetries}
that minimizes only the local pairwise differences between the predicted actions across the trajectory.
As the Action-Angle Network is completely differentiable,
the parameters $\phi$ can be obtained via
gradient-descent-based minimization of the total loss:
$
    \phi^* = \text{argmin}_\phi \left(L_{\text{predict}} + \lambda L_{\text{action}}\right),
$
where $\lambda$ is a hyperparameter that controls the strength of the regularization.

\textbf{Schedule for $\Delta t$}:
Increasing $\Delta t$ over the course of training helped,
as it corresponded to gradually increasing the complexity of the prediction task.
We set $\Delta t_{\text{max}} = 10$ and sampled $\Delta t$ according to:
\begin{align}
    \Delta t \sim \text{Uniform}\left(0, (\text{Current training step})/(\text{Maximum training steps})\right) \cdot \Delta t_{\text{max}}
\end{align}
\vspace{-2em}


\section{Experiments}
\label{sec:experiments}

We compare the Action-Angle Network to three strong baseline models: the Euler Update Network (EUN), the Neural Ordinary Differential Equations (Neural ODE) \citep{chen:node}, and the physics-inspired Hamiltonian Neural Networks (HNN)
\citep{hnn}. A comparison of these models can be found in \autoref{tab:model_comparison}.
The baseline models are further described in \autoref{sec:baselines}.

\begin{table*}[htbp]
    \centering
    \caption{Comparing different models.}
    \label{tab:model_comparison}
    \rowcolors{2}{}{gray!10}
    \begin{tabular}{*5c}
        \toprule
                                  & Action-Angle Network  & EUN  & Neural ODE     & HNN             \\    
        \midrule
        Parameter count           & $\approx 8.5$K        & $\approx 9$K         & $\approx 100$K & $\approx 200$K   \\
        Inference time            & $O(1)$                & $O(1)$                & $O(\Delta t)$  & $O(\Delta t)$    \\
        Learns conservation laws  & \checkmark            &                       &                & \checkmark       \\
        Learns linear dynamics    & \checkmark            &  \checkmark           &                &                  \\
        \bottomrule
    \end{tabular}
\end{table*}

\begin{figure}[htbp]
     \centering
     \begin{tabular}[c]{cc}
     \begin{subfigure}[c]{0.38\textwidth}
         \centering
         \includegraphics[width=\textwidth]{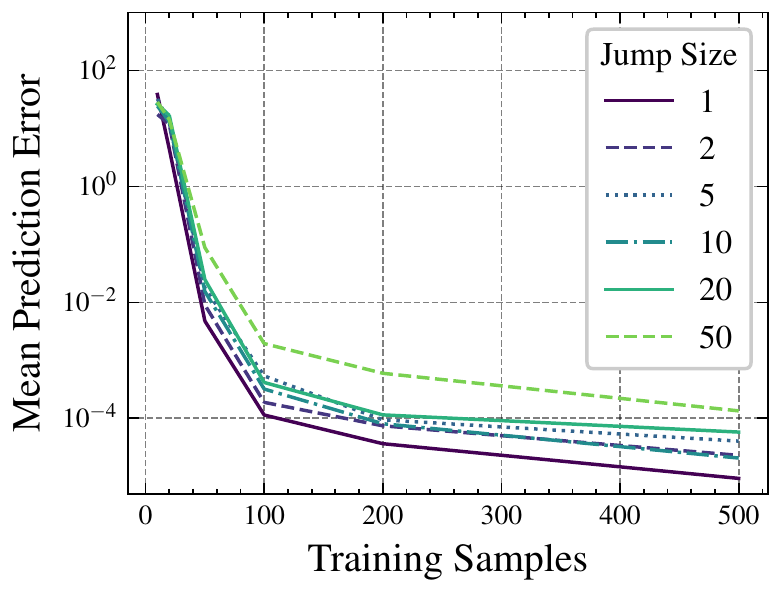}
         \caption{Action-Angle Network}
         \label{fig:action_angle_prediction_error}
     \end{subfigure}
     \hfill
     \begin{subfigure}[c]{0.38\textwidth}
         \centering
         \includegraphics[width=\textwidth]{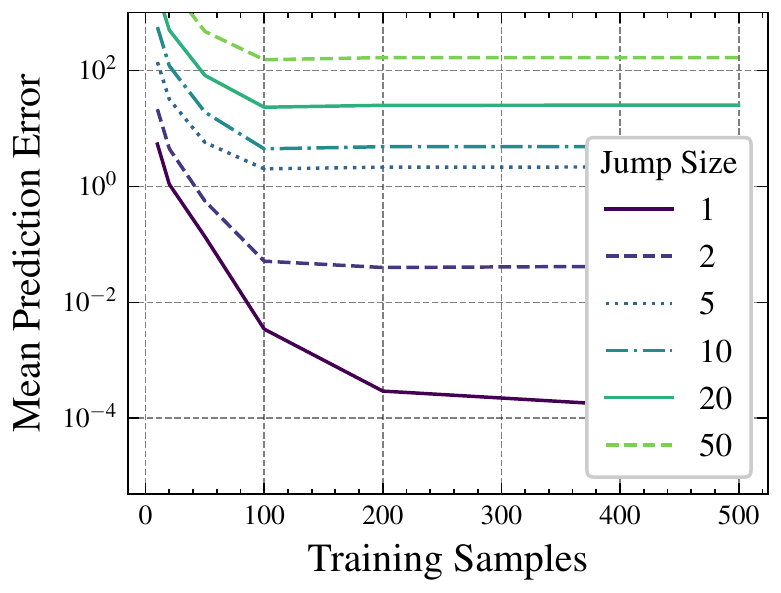}
         \caption{Euler Update Network}
         \label{fig:euler_update_prediction_error}
     \end{subfigure}
     \end{tabular}
     \\
     \begin{tabular}[c]{cc}
     \begin{subfigure}[b]{0.38\textwidth}
         \centering
         \includegraphics[width=\textwidth]{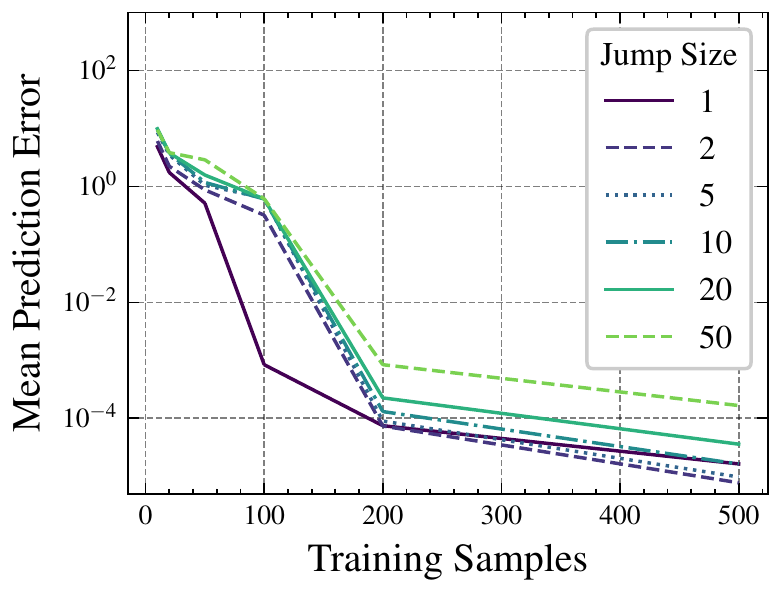}
         \caption{Neural ODE}
         \label{fig:neural_ode_prediction_error}
     \end{subfigure}
          \hfill
     \begin{subfigure}[b]{0.38\textwidth}
         \centering
         \includegraphics[width=\textwidth]{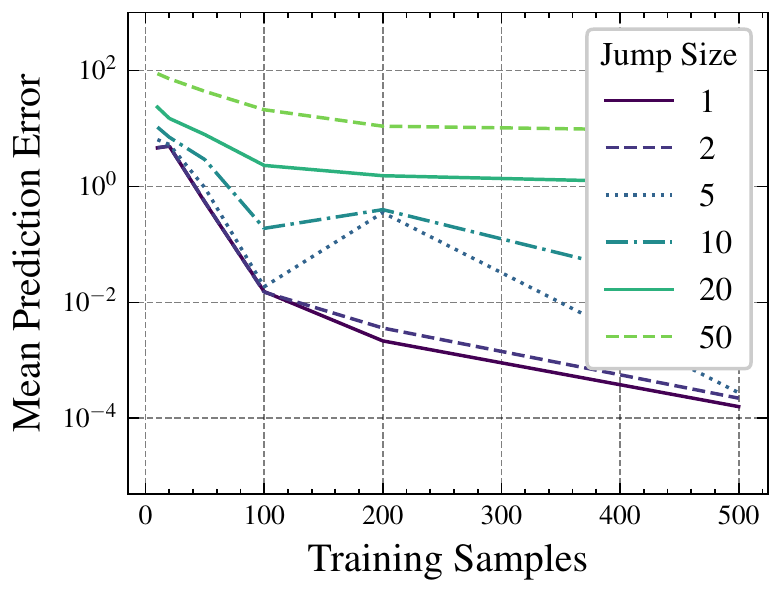}
         \caption{Hamiltonian Neural Network}
         \label{fig:hamiltonian_network_prediction_error}
     \end{subfigure}
     \end{tabular}
     \caption{Prediction errors on test data as a function of training samples.}
     \label{fig:performance_vs_samples}
\end{figure}

\begin{figure}[htbp]
    \centering
    \begin{tabular}[c]{cc}
    \begin{subfigure}[b]{0.45\textwidth}
        \centering
        \includegraphics[width=\textwidth]{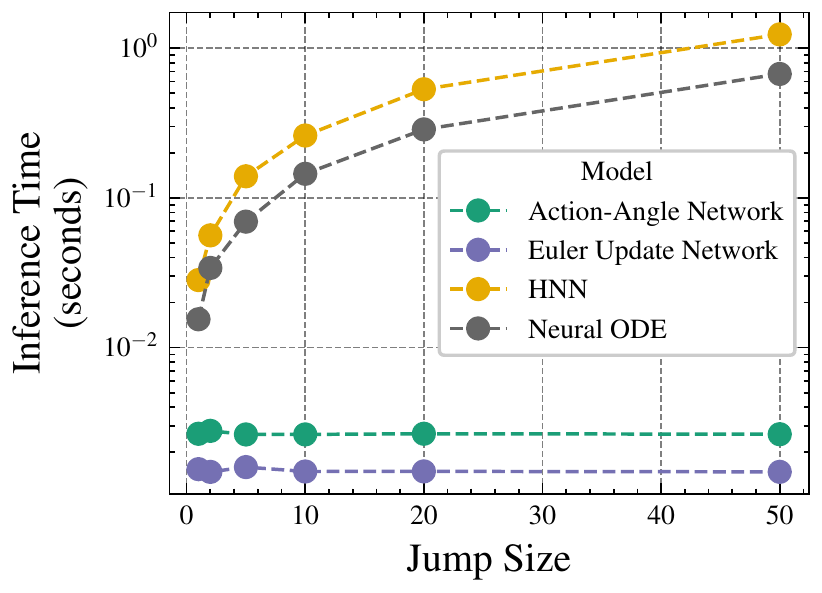}
        \caption{Inference times as a function of jump $\Delta t$.}
        \label{fig:inference_times}
    \end{subfigure}
    \hfill
    \begin{subfigure}[b]{0.42\textwidth}
        \centering
        \includegraphics[width=\textwidth]{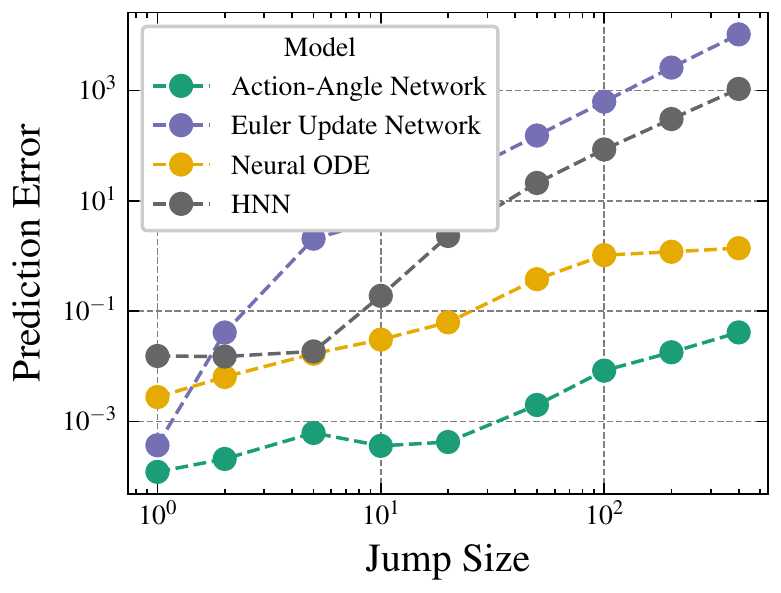}
        \caption{Prediction error as a function of jump $\Delta t$.}
        \label{fig:performance_vs_time}
    \end{subfigure}
    \end{tabular}
    \caption{Evaluating each model on (a) inference time and (b) prediction error.}
    \label{fig:evaluation}
\end{figure}

\begin{figure}[!htbp]
    \centering
    \begin{tabular}[c]{cc}
    \begin{subfigure}[c]{0.4\textwidth}
        \centering
        \includegraphics[width=\textwidth]{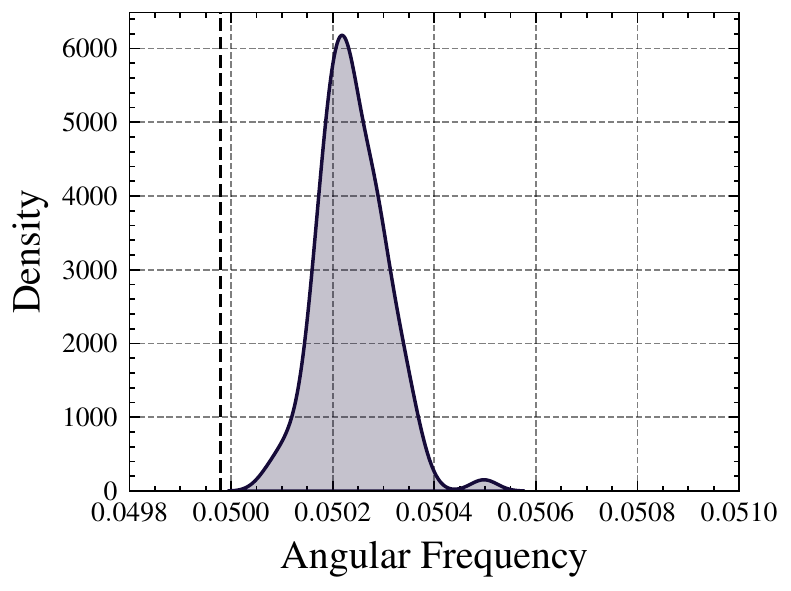}
    \end{subfigure}
    \hfill
    \begin{subfigure}[c]{0.4\textwidth}
        \centering
        \includegraphics[width=\textwidth]{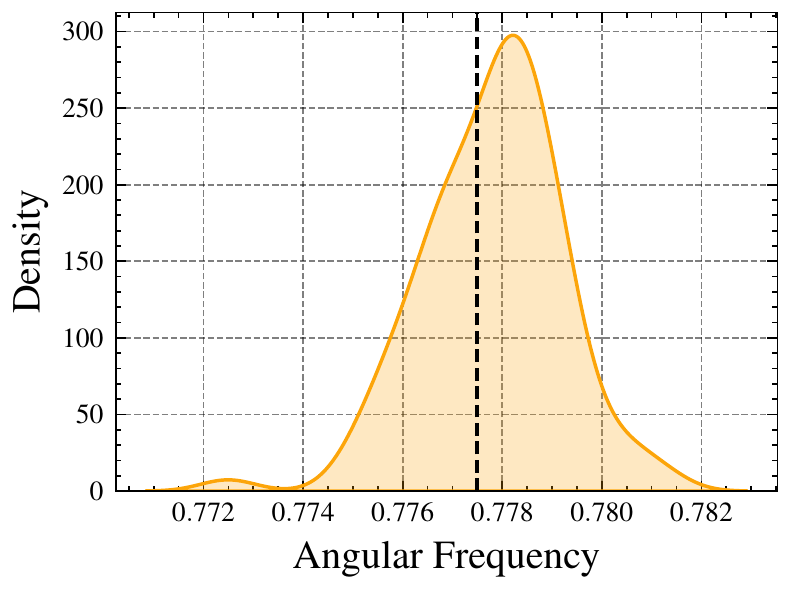}
    \end{subfigure}
    \end{tabular}
    \caption{Kernel density plot of angular frequencies predicted by the Action-Angle Network. True angular frequencies are indicated by dashed vertical lines.}
    \label{fig:angular_frequencies}
\end{figure}

As detailed in \autoref{sec:harmonic},
we simulate a system of coupled harmonic oscillators.\footnote{We have created animations of the system trajectory and model predictions at \href{https://ameya98.github.io/ActionAngleNetworks/webpage/index.html}{this webpage}. Our codebase to run all experiments and analyses is available at \url{https://github.com/ameya98/ActionAngleNetworks}.} \autoref{fig:performance_vs_samples} depicts the prediction error for each of the models as a function of training samples, showing that the Action-Angle Network is much more data-efficient than the other baselines. \autoref{fig:inference_times} shows that the Action-Angle Network can be queried much faster than the Neural ODE and the HNN, with an inference time that is independent of $\Delta t$. \autoref{fig:performance_vs_time} depicts the prediction error as a function of $\Delta t$, showing that the Action-Angle Network also scales much better with the jump size $\Delta t$, even for jump sizes larger than those seen during training $\Delta t_{\max} = 10$. Finally, \autoref{fig:angular_frequencies} shows that the predicted angular frequencies from the Action-Angle Network closely match the true angular frequencies.

\section{Conclusion}

Our preliminary experiments indicate that Action-Angle Networks can be a promising alternative for
learning efficient physical simulators.
Action-Angle Networks overcome many of the obstacles -- instability and inefficiency -- faced by
state-of-the-art learned simulators today.
That being said, our
experiments here only model very simple integrable systems.
Next,
we plan to analyse the performance of Action-Angle
Networks on modelling large-scale non-integrable systems 
(such as cosmological simulations and molecular dynamics trajectories) to better understand their tradeoffs.


\section*{Broader Impact Statement}

In this paper, we have proposed a novel method for modelling the dynamics of Hamiltonian systems, with improvements
in computational efficiency.
There has been much interest in accelerating simulations of various kinds 
across scientific domains that could have impact on human lives (for 
example, simulating biological cell cycles and immune system responses).
However, given the techniques here apply to highly-constrained physical systems, we do not anticipate any negative social implications of our
work.


\begin{ack}
    The authors acknowledge the MIT SuperCloud and Lincoln Laboratory Supercomputing Center for providing (HPC, database, consultation) resources that have contributed to the research results reported within this paper.
\end{ack}


\bibliographystyle{plainnat}
\bibliography{references}


\section*{Checklist}

\begin{enumerate}

\item For all authors...
\begin{enumerate}
    \item Do the main claims made in the abstract and introduction accurately reflect the paper's contributions and     scope?
        \answerYes{We have designed and evaluated Action-Angle Networks, showing how they can efficiently learn the dynamics of integrable systems.}
    \item Did you describe the limitations of your work?
        \answerYes{Action-Angle Networks can be applied to only integrable systems. Their performance on non-integrable Hamiltonian systems still needs investigation.}
    \item Did you discuss any potential negative societal impacts of your work?
        \answerYes{We have mentioned an overview in the Broader Impact statement. We do not expect any negative societal impacts of our work. }
    \item Have you read the ethics review guidelines and ensured that your paper conforms to them?
        \answerYes{Our paper does not utilise or expose any human-derived data. All data for the physical systems discussed here are synthetically generated via numerical integration.}
\end{enumerate}

\item If you are including theoretical results...
\begin{enumerate}
    \item Did you state the full set of assumptions of all theoretical results?
        \answerNA{}
    \item Did you include complete proofs of all theoretical results?
        \answerNA{}
\end{enumerate}

\item If you ran experiments...
\begin{enumerate}
    \item Did you include the code, data, and instructions needed to reproduce the main experimental results (either in the supplemental material or as a URL)?
        \answerYes{Yes, we have added a link to our code in \autoref{sec:experiments}.}
    \item Did you specify all the training details (e.g., data splits, hyperparameters, how they were chosen)?
        \answerYes{Yes, the entire training pipeline and configuration is available at the link for the code in \autoref{sec:experiments}.}
    \item Did you report error bars (e.g., with respect to the random seed after running experiments multiple times)?
        \answerNo{We plan to address this in a future version.}
    \item Did you include the total amount of compute and the type of resources used (e.g., type of GPUs, internal cluster, or cloud provider)?
        \answerYes{Experiments were run on the CPU nodes of MIT SuperCloud, and can be easily reproduced with minimal resources.}
\end{enumerate}

\item If you are using existing assets (e.g., code, data, models) or curating/releasing new assets...
\begin{enumerate}
    \item If your work uses existing assets, did you cite the creators?
        \answerNA{}
    \item Did you mention the license of the assets?
        \answerYes{Our code linked in \autoref{sec:experiments} is licensed under the Apache License 2.0.}
    \item Did you include any new assets either in the supplemental material or as a URL?
        \answerYes{We have released code and simulation tools at the link in \autoref{sec:experiments}.}
    \item Did you discuss whether and how consent was obtained from people whose data you're using/curating?
        \answerNA{}
    \item Did you discuss whether the data you are using/curating contains personally identifiable information or offensive content?
        \answerNA{}
\end{enumerate}

\item If you used crowdsourcing or conducted research with human subjects...
\begin{enumerate}
    \item Did you include the full text of instructions given to participants and screenshots, if applicable?
        \answerNA{}
    \item Did you describe any potential participant risks, with links to Institutional Review Board (IRB) approvals, if applicable?
        \answerNA{}
    \item Did you include the estimated hourly wage paid to participants and the total amount spent on participant compensation?
        \answerNA{}
\end{enumerate}

\end{enumerate}


\appendix
\section{Appendix}

\subsection{Baseline Models}
\label{sec:baselines}
\textbf{Euler Update Network}: This model updates the latent state via the forward Euler method:
\begin{align*}
    \hat{u}(t) &\gets \text{Encode}(u(t)) \\
    \hat{u}(t + \Delta t) &\gets \hat{u}(t) + \Delta t \cdot \gF(\hat{u}(t)) \\
    u(t + \Delta t) &\gets \text{Decode}(\hat{u}(t + \Delta t)) 
\end{align*}

\textbf{Neural Ordinary Differential Equations}: A generalization of 
Euler Update Networks, Neural ODEs \citep{chen:node} have demonstrated state-of-the-art performance on several time-series forecasting problems. They usually use a higher-order numerical integration scheme
to update the latent state:
\vspace{-0.4em}
\begin{align*}
    \hat{u}(t) &\gets \text{Encode}(u(t))  \\
    \hat{u}(t + \Delta t) &\gets \hat{u}(t) + {\int_{t}^{t + \Delta t}}\gF(\hat{u}(s)) ds \\ 
    u(t + \Delta t) &\gets \text{Decode}(\hat{u}(t + \Delta t)) 
\end{align*}

\textbf{Hamiltonian Neural Network}: Hamiltonian Neural Networks (HNNs) \citep{hnn} models the Hamiltonian in a latent space explicitly and updates the latent coordinates via Hamilton's equations.
\begin{align*}
    \hat{q}(t), \hat{p}(t) &\gets \text{Encode}(u(t))  \\
    \hat{q}(t + \Delta t) &\gets \hat{q}(t) + {\int_{t}^{t + \Delta t}} \pdd{}{\hat{p}}H(\hat{q}(s), \hat{p}(s)) ds \\
    \hat{p}(t + \Delta t) &\gets \hat{p}(t) - {\int_{t}^{t + \Delta t}} \pdd{}{\hat{q}}H(\hat{q}(s), \hat{p}(s)) ds \\
    u(t + \Delta t) &\gets \text{Decode}(\hat{q}(t + \Delta t), \hat{p}(t + \Delta t)) 
\end{align*}

Wherever applicable, we used the Dormand-Prince 5(4) solver, a 5th order Runge-Kutta method for numerical integration.

All of these baselines can technically simulate the Action-Angle Network by encoding the canonical coordinates into the action-angle coordinates to linearize the dynamics. 

\subsection{Harmonic Oscillators}
\label{sec:harmonic}

We model a system of $n$ point masses, which are connected to a wall
via springs of constant $k_w$ and to each other via springs of constant $k_p$.
This can be described by the set of $n$ differential equations:
\begin{align}
    \label{eqn:harmonic}
    m \frac{d^2q_i}{dt^2} = -k_w q_i + {\sum_{j \neq i}} k_p (q_j - q_i).
\end{align}
where $q_i(t)$ is the position of the $i$th point mass.
This forms a system of coupled harmonic oscillators, where $k_p$ controls the strength of the coupling.
Any solution to \autoref{eqn:harmonic} is a linear combination of the `normal modes' of the system \citep{morin:normal-modes}:
\begin{align*}
    q(t) = {\textstyle \sum_{r = 1}^n} A_r c_r \cos{(\omega_r t + \phi_r)}.
\end{align*}
The $n$ angular frequencies $\omega$ and coefficients $c$ are found by solving the following eigenvalue-eigenvector equation $(M + \omega^2 \mathbb{I}_n) c = 0$, 
where $M$ is the matrix defined as:
\begin{align}
    M_{ij} =
    \begin{cases}
        -\dfrac{k_w + (n - 1)k_p}{m_i} \ &\text{if} \ i = j \\
        \dfrac{k_p}{m_i} \ &\text{if} \ i \neq j
    \end{cases}
\end{align}


\end{document}